# A Deep Learning Framework Integrating CNN and BiLSTM for Financial Systemic Risk Analysis and Prediction

Yu Cheng
Columbia University
New York, USA

Zhen Xu
Independent Researcher
Shanghai, China

Yuan Chen
Rice University
Houston, USA

Yuhan Wang
Columbia University
New York, USA

Zhenghao Lin
Northeastern University
Boston, USA

Jinsong Liu*
University at Buffalo
Buffalo, USA

*Abstract*-This study proposes a deep learning model based on the combination of convolutional neural network (CNN) and bidirectional long short-term memory network (BiLSTM) for discriminant analysis of financial systemic risk. The model first uses CNN to extract local patterns of multidimensional features of financial markets, and then models the bidirectional dependency of time series through BiLSTM, to comprehensively characterize the changing laws of systemic risk in spatial features and temporal dynamics. The experiment is based on real financial data sets. The results show that the model is significantly superior to traditional single models (such as BiLSTM, CNN, Transformer, and TCN) in terms of accuracy, recall, and F1 score. The F1-score reaches 0.88, showing extremely high discriminant ability. This shows that the joint strategy of combining CNN and BiLSTM can not only fully capture the complex patterns of market data but also effectively deal with the long-term dependency problem in time series data. In addition, this study also explores the robustness of the model in dealing with data noise and processing high-dimensional data, providing strong support for intelligent financial risk management. In the future, the research will further optimize the model structure, introduce methods such as reinforcement learning and multimodal data analysis, and improve the efficiency and generalization ability of the model to cope with a more complex financial environment.

*Keywords-Financial systemic risk, convolutional neural network, bidirectional long short-term memory network, deep learning*

## I.  Introduction

Managing and controlling financial systemic risk is an important research direction in the financial field. With the increasing complexity of market structure and the high correlation of the global economy, traditional risk management methods face many challenges in dealing with nonlinear characteristics, time series dependencies, and high-dimensional data processing. Among emerging research areas, deep learning-based models for risk identification have gained significant attention. Notably, bidirectional long short-term memory networks and convolutional neural networks are frequently applied in financial risk management due to their excellent capabilities in time series analysis and feature extraction [1].

Systemic risks in financial markets are often the result of the combined effects of multiple economic variables and are affected by a combination of factors such as market fluctuations, interest rate changes, macroeconomic environment, and policy regulation [2]. Therefore, fully mining and utilizing multidimensional data to build a risk management model with real-time and accurate identification capabilities has become an important research direction. In this context, deep learning models, with their powerful nonlinear mapping and high-dimensional feature extraction capabilities, have shown significant advantages over traditional statistical methods and can effectively improve the accuracy and robustness of risk identification [3].

The bidirectional long short-term memory network is a deep learning architecture designed to learn dependencies in both forward and backward directions within time series data. It can extract potential features from both forward and backward sequences and is particularly suitable for processing financial time series data [4]. In systemic risk prediction, price fluctuations, trading volume changes, and macroeconomic indicators usually exhibit complex temporal dependencies. BiLSTM effectively alleviates the information-forgetting problem of traditional models in long-term dependency modeling through its unique memory unit and gating mechanism, thereby providing more stable and timely decision support for financial risk prediction [5].

Convolutional neural network (CNN) was originally used for image processing, but its advantages in one-dimensional feature extraction have gradually attracted attention [6-7]. CNN can efficiently extract key patterns and local features in data through local perception and weight-sharing mechanisms. When the time series data of the financial market is processed through data enhancement and multi-scale feature extraction technology [8], CNN can identify potential market risk signals, thereby providing strong support for early warning of systemic risks. In addition, the hierarchical structure of CNN can effectively realize multi-level feature fusion, further improving the performance of the model.

In this study, we jointly modeled BiLSTM and CNN to build a deep learning framework that integrates time series modeling and feature extraction capabilities. The framework first uses BiLSTM to extract time series features from financial market data and then performs multi-level convolution operations on these features through CNN to extract more refined and in-depth features. Through this multi-model collaborative design, the accuracy and robustness of systemic risk identification can be significantly improved to meet the needs of real-time risk prediction in complex financial environments.

## II. RELATED WORK

The integration of convolutional neural networks (CNNs) and bidirectional long short-term memory networks (BiLSTMs) in deep learning models has demonstrated significant potential in addressing complex predictive tasks, particularly those involving time series and high-dimensional data. Previous studies have laid a strong foundation by exploring key methodologies that contribute to the improvement of accuracy, robustness, and efficiency in predictive models.

CNNs have proven effective in feature extraction through their ability to identify local patterns within large datasets, which is essential for capturing key signals in multidimensional data. Du [9] demonstrated optimized CNN architectures, highlighting their importance in identifying subtle patterns while maintaining computational efficiency, a critical consideration for real-time applications. Wu et al. [10] emphasized the role of CNNs in enhancing feature representations through hierarchical structures, which is directly applicable to this work's design for capturing both local and global risk indicators. Additionally, Liang et al. [11] showcased the utility of autoencoders in dimensionality reduction and feature extraction, which complements CNN-based models by reducing noise and enhancing the discriminative power of extracted features in high-dimensional settings.

BiLSTM architectures have gained attention for their ability to model sequential dependencies in both forward and backward directions, addressing the common issue of long-term dependency in time series data. Yao [12] highlighted how time-series reinforcement learning frameworks benefit from models capable of learning complex temporal patterns, which supports the role of BiLSTM in this study for modeling financial market dependencies. Huang and Yang [13] addressed the challenge of feature redundancy in time series models, proposing strategies that align with the goal of this research in avoiding overfitting and improving generalization when dealing with long time horizons.

The synergy between different deep learning techniques is an area of active research, with several studies demonstrating that hybrid architectures outperform standalone models. Long et al. [14] explored adaptive neural network frameworks to dynamically adjust to evolving data characteristics, a concept that aligns with the goal of improving robustness under noisy and volatile conditions. Yan et al. [15] highlighted the use of neural architecture search to optimize hybrid model structures, providing insights into how this work could further refine its design for better performance and computational efficiency. Efficient feature selection and dimensionality reduction are also key to enhancing model performance in high-dimensional environments. Hu et al. [16] proposed adaptive weighting techniques for handling sparse and imbalanced datasets, which can be applied to improve model robustness in scenarios where certain financial indicators may dominate or vary significantly over time.

Emerging techniques, such as graph-based models, are also providing valuable perspectives for designing more effective risk management systems. Yao et al. [17] introduced hierarchical graph neural networks, which suggest potential avenues for extending this work by capturing relational dependencies among financial entities and integrating them with temporal models.

Overall, these studies collectively contribute to the development of the proposed CNN-BiLSTM framework by providing insights into effective feature extraction, sequential modeling, dimensionality reduction, and robustness under dynamic conditions. The lessons drawn from hybrid architectures, automated feature selection, and optimization strategies play a crucial role in guiding the model design, aiming for superior performance in the identification and prediction of financial systemic risk.

## III. METHOD

This research introduces a deep learning framework combining convolutional neural networks and bidirectional long short-term memory networks for analyzing financial systemic risk. The model first extracts the local features of financial time series data through CNN and then uses BiLSTM to model its long-term dependencies, thereby comprehensively characterizing the dynamic changes of financial data in feature space and time domain, and improving the accuracy and robustness of risk prediction. The model structure is shown in Figure 1.

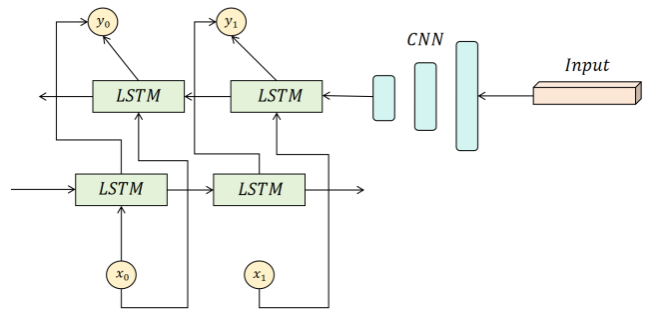

Figure 1 Overall model architecture

At the model input stage, we represent the time series data of the financial market as a multi-dimensional feature matrix, denoted as $X \in R^{T \times F}$, where $T$ represents the time step and

$F$ represents the number of features. These features can include stock prices, trading volumes, interest rates, macroeconomic indicators, etc. The input data is first normalized and standardized to ensure that the scale differences of data in different dimensions do not affect the learning effect of the model.

First, a convolutional neural network (CNN) is used to extract local features from the input data. For each time step, multiple one-dimensional convolution kernels are applied to extract features of different scales through convolution operations. The formula for the convolution operation is:

$$h_t^c = \text{Re}LU(W_c * X_t + b_c)$$

Among them, $W_c$ represents the convolution kernel, $*$ represents the convolution operation, $X_t$ is a time step of the input data, $b_c$ is the bias term, and $h_t^c$ is the output after convolution. The activation function uses ReLU to introduce nonlinear features. Moreover, the model employs a max-pooling layer to decrease feature dimensions, enhancing computational efficiency and boosting its generalization performance. After multiple layers of convolution and pooling operations, the extracted feature matrix is input into the BiLSTM layer.

Next, the bidirectional long short-term memory network (BiLSTM) is used to model the forward and backward correlation of the time series from the features extracted by CNN. BiLSTM realizes bidirectional modeling of the time series by combining the forward and backward LSTM layers. The hidden state of the output is expressed as:

$$h_t = \overrightarrow{h_t} + \overleftarrow{h_t}$$

Among them, $\overrightarrow{h_t}$ represents the hidden state of the forward LSTM, and $\overleftarrow{h_t}$ represents the hidden state of the backward LSTM. Through this bidirectional structure, the model can make full use of past and future information to improve the ability to identify systemic risks. In order to further improve the robustness of the model, the attention mechanism is applied to the output layer of BiLSTM to calculate the importance weight of each time step and perform a weighted average on the output [18].

Finally, the model maps the output of BiLSTM to the final risk discrimination result through a fully connected layer, which is expressed as:

$$y = \sigma(W_o h_T + b_o)$$

Among them, $W_o$ represents the weight matrix of the output layer, $h_T$ is the output of BiLSTM at the last time step, $b_o$ is the bias term, and $\sigma(\cdot)$ represents the Sigmoid activation function, which is used to generate the probability value of risk discrimination. The objective function of the model is the cross-entropy loss, which is optimized through back propagation and gradient descent algorithms to minimize the error between the predicted results and the true labels.

In summary, the model extracts local time series features through CNN and then uses BiLSTM to model time series, successfully achieving a comprehensive characterization of financial systemic risks. The model has excellent learning capabilities in multiple feature dimensions and time series dependencies, can effectively cope with the complex dynamic changes in the financial market, and significantly improve the prediction accuracy of systemic risks.

IV. EXPERIMENT

A. Datasets

The financial systemic risk identification dataset used in this study comes from public real financial market data sources such as Yahoo Finance. The dataset covers historical transaction data of major global financial markets, including key indicators such as daily closing prices, trading volumes, volatility, and interest rate levels of financial instruments such as stocks, bonds, foreign exchange, and futures. These data are obtained through API interfaces, with a long period and covering multiple economic cycles. They can reflect the performance of the market under different economic environments and are an ideal data basis for building financial risk identification models.

In addition, in order to improve the generalization ability and practical application value of the model, the dataset has been screened and cleaned at multiple levels to eliminate data missing and abnormal volatility samples with the method proposed by Li et al. [19]. At the same time, through the time window division method, the continuous financial market data is divided into a training set, validation set, and test set. This division method can not only ensure the effective learning of the model on historical data but also provide a reference for risk warning in future market changes. The real source and high-quality characteristics of the data set provide solid data support for the research and model construction of financial systemic risks.

B. Experimental Results

This research conducts comparative experiments using multiple deep learning models to validate the effectiveness of the proposed CNN-BiLSTM framework in detecting financial systemic risk. First, the traditional bidirectional long short-term memory network (BiLSTM) [20] is used. This model performs well in time series data modeling but has limitations in feature extraction. Secondly, the convolutional neural network (CNN) [21] is used, which has obvious advantages in local feature extraction but cannot model time dependencies. In addition, the Transformer-based time series classification model (Transformer) is introduced. This model models global time dependencies through the attention mechanism and demonstrates powerful feature expression capabilities. Finally, the hybrid structure of the temporal convolutional network (TCN) [22] is used. This model effectively models long time series through dilated convolution and residual structure. Under the same dataset and experimental settings, the above models are fully trained and tested to comprehensively evaluate their performance in the task of systemic risk identification. The experimental results are shown in Table 1.

Table 1 Experimental results

| Model | ACC | Recall | F1-score |
|---|---|---|---|
| BILSTM | 0.84 | 0.82 | 0.83 |
| CNN | 0.81 | 0.79 | 0.80 |
| Transformer | 0.86 | 0.85 | 0.85 |
| TCN | 0.85 | 0.83 | 0.84 |
| BILSTM+CNN(Ours) | 0.89 | 0.87 | 0.88 |

The experimental findings reveal notable variations in the performance of different deep learning models when detecting financial systemic risks. Single models such as BiLSTM and CNN show certain advantages in their respective areas of expertise. BiLSTM has an accuracy of 0.84, a recall of 0.82, and an F1-score of 0.83, indicating that it has excellent capabilities in time series modeling and can effectively capture the bidirectional dependencies of data. However, due to the lack of a spatial feature extraction mechanism, its ability to model complex features is limited, resulting in overall prediction performance that is slightly inferior to that of a combined model with a more complex structure.

The CNN model performed slightly lower in the experiment, with an accuracy of 0.81, a recall of 0.79, and an F1-score of 0.80. This result shows that although CNN performs well in local feature extraction, when processing time series data in the financial market, it cannot capture global time dependencies, which limits the performance of the model in long-term trend prediction and time series dynamic changes. Therefore, the CNN model performs slightly poorly in the task of identifying financial systemic risks, especially in scenarios where long-term historical data is required, and where its feature extraction capability is relatively weak.

The time series classification model based on the Transformer exhibits excellent performance, attaining an accuracy of 0.86, a recall of 0.85, and an F1-score of 0.85. This is mainly due to the built-in self-attention mechanism of the Transformer model, which can capture the global time dependency and complex nonlinear relationships in the data. In the time series data processing of the financial market, the Transformer model can dynamically focus on important time steps and extract key features, thereby achieving more efficient risk identification. However, the Transformer model still has room for optimization in high-frequency data processing and local detail modeling.

The TCN model also performs well, achieving a precision of 0.85, a recall of 0.83, and an F1 score of 0.84. Its architecture uses dilated convolutions and residual connections for efficient long-term series modeling and performs well in the task of identifying systemic financial risks. However, TCN relies on a lot of data preprocessing, which poses challenges in handling multi-dimensional data fusion and nonlinear feature representation. These findings suggest that while TCN performs well in various time series tasks, it may struggle to provide optimal performance when applied independently in complex financial scenarios.

Finally, the proposed BiLSTM+CNN model performed best in all indicators, with an accuracy of 0.89, a recall of 0.87, and an F1-score of 0.88. This result proves the effectiveness of the joint strategy of bidirectional time series modeling and local feature extraction. In the model design, CNN first extracts local features to capture short-term dynamic changes in market data, while BiLSTM further models the long-term dependence of data and extracts important relationships between the previous and next sequences. This deep combination of feature extraction and time series modeling not only improves the model's sensitivity to systemic risks but also enhances the model's overall predictive ability. The experimental results fully demonstrate the robustness and wide applicability of the model, providing strong support for the intelligent management and risk warning of financial systemic risks.

Furthermore, we give a graph of the loss function drop, as shown in Figure 2.

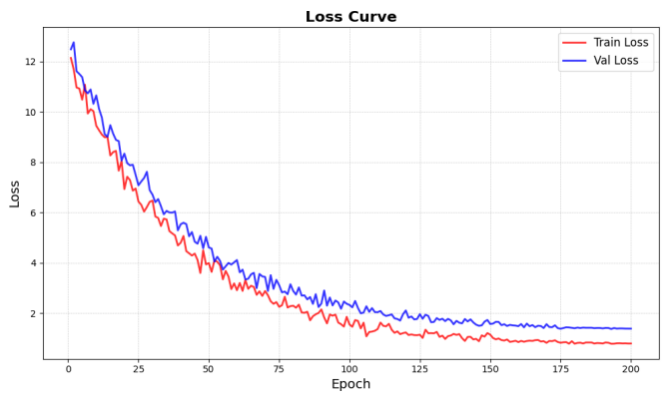

Figure 2. Loss function changes with epoch

As shown in the figure, the training loss and validation loss decrease progressively during the training process, indicating continuous model optimization and improved fitting to both training and validation data patterns. This trend shows that the learning process of the model is effective, the parameters are gradually adjusted during training, the loss function value is reduced, and the performance of the model is gradually improved.

A noticeable gap exists between the training loss and validation loss, particularly in the later training stages. The training loss is markedly lower than the validation loss, suggesting that the model fits the training data better than the validation data, potentially indicating some level of overfitting. Nevertheless, the downward curve of the validation loss shows that the model still has a certain generalization ability on the validation set, which can be further optimized by introducing regularization or early stopping strategies.

In the later stages of training, the validation loss curve stabilizes, indicating that the model is approaching convergence. Continuing to increase the number of training epochs may not significantly improve performance and could raise the risk of overfitting. Therefore, the current training setup yields relatively satisfactory results, though attention should still be given to the gap between validation and training losses to enhance the model's generalization ability.

## V. Conclusion

This study introduces a deep learning model that integrates CNN and BiLSTM for financial systemic risk analysis. Experimental results demonstrate its superiority over traditional single models such as BiLSTM, CNN, Transformer, and TCN in terms of accuracy, recall, and F1 score. The findings confirm the effectiveness of combining convolutional feature extraction with bidirectional time-series modeling in complex financial environments. The proposed model excels at capturing multidimensional features and temporal dependencies in financial markets, offering robust technical support for intelligent risk management.

Although the research has achieved remarkable results, there are also certain limitations. For example, the performance of the model may be limited by the quality and scale of the data set, and its generalization ability needs to be further verified when facing more volatile and non-stationary financial markets. In addition, the computational complexity of the model is high, and the training and prediction processes require large computing resources, which may bring challenges in application scenarios with high real-time requirements. Therefore, future research can try to further improve the efficiency and practicality of the model through methods such as model optimization, feature selection, and multimodal data fusion.

Looking ahead, the prediction and management of financial market risks will still be a research direction full of opportunities and challenges. With the improvement of data availability and the continuous enhancement of computing power, deep learning technology will show broader application prospects in the financial field. Future research can combine emerging algorithms such as reinforcement learning and contrastive learning to explore more intelligent risk control and decision support models. In addition, combining the model with blockchain technology for distributed financial risk management is also a direction worth exploring. These innovative studies will promote the development of financial technology and help the stability and healthy operation of the global financial market.